\documentclass{article}

\PassOptionsToPackage{numbers,compress}{natbib}
    
\usepackage[final]{neurips_2019}

\usepackage[utf8]{inputenc} 
\usepackage[T1]{fontenc}    
\usepackage[colorlinks=true,linkcolor=blue,citecolor=blue]{hyperref}       
\usepackage{url}            
\usepackage{booktabs}       
\usepackage{amsfonts}       
\usepackage{nicefrac}       
\usepackage{microtype}      
\usepackage{graphicx}
\usepackage{multirow}
\usepackage{amsmath}
\usepackage{subcaption}

\title{Training Compact Models for Low Resource Entity Tagging using Pre-trained Language Models}

\author{%
  Peter Izsak, Shira Guskin, Moshe Wasserblat\\
  Intel AI Lab\\
  \texttt{\{peter.izsak, shira.guskin, moshe.wasserblat\}@intel.com} \\
}
\begin{document}
\maketitle

\begin{abstract}
    Training models on low-resource named entity recognition tasks has been shown to be a challenge \cite{Zhang2016NameTF},
    especially in industrial applications where deploying updated models is a continuous effort and crucial for business operations.
    In such cases there is often an abundance of unlabeled data, while labeled data is scarce or unavailable. 
    Pre-trained language models trained to extract contextual features from text were shown to improve many natural language processing (NLP) tasks, including scarcely labeled tasks, by leveraging transfer learning.
    However, such models impose a heavy memory and computational burden, making it a challenge to train and deploy such models for inference use. In this \textit{work-in-progress} we combined the effectiveness of transfer learning provided by pre-trained masked language models with a semi-supervised approach to train a fast and compact model using labeled and unlabeled examples. Preliminary evaluations show that the compact models can achieve competitive accuracy with 36$\times$ compression rate when compared with a state-of-the-art pre-trained language model, and run significantly faster in inference, allowing deployment of such models in production environments or on edge devices.
\end{abstract}

\section{Introduction}

Named Entity Recognition (NER) is a core NLP task in which a system classifies words or phrases using a predefined set of labels. NER is mostly used in Information Extraction use cases and is widely deployed in industrial applications. In recent years, many deep learning approaches have been proposed \cite{Ma2016EndtoendSL, Collobert2011NaturalLP, Lample2016NeuralAF, Chiu2015NamedER, Strubell2017FastAA} which proved to be more successful in learning word, phrase and character features and entity classification. However, in cases when there is an abundance of data of which none or very little is annotated, the performance of such models is substantially degraded since standard approaches fail to learn good feature representation from small amounts of labeled examples.
Annotating more data could overcome poor model performance, however, it requires expert knowledge and monetary resources and may be bound by time-related constraints.

Recently, many NLP tasks have shown great improvements thanks to pre-trained language models. Models such as ELMo \cite{Peters2018DeepCW}, BERT \cite{Devlin2018BERTPO} and GPT \cite{Radford2018ImprovingLU} include a large language model or contextual embedder with hundreds of millions of parameters trained on large datasets and were shown to produce state-of-the-art models for many NLP tasks, including low-resource scenarios where very little annotated data is available.
In particular, Transformer-base \cite{Vaswani2017AttentionIA} models such as BERT were shown to be highly effective in transfer-learning by fine-tuning all model parameters when training on a target task. However, the increase in model accuracy comes at a price of increased memory and computation requirements and with high latency when used in inference. Therefore, deploying such models in production environments is difficult or completely impossible on edge devices.

In this preliminary work we strive to train a fast and compact model, that is, a model with competitive accuracy in low-resource NER tasks compared with a pre-trained language model, which is also small enough to fit on edge devices or be more energy efficient for production deployment.
For this purpose we employed two methods: model distillation \cite{Hinton2015DistillingTK} to train a compact \textit{student} network using a much larger \textit{teacher} network, and pseudo-labelling \cite{Lee2013PseudoLabelT} to produce \textit{pseudo} guesses by the large network for utilizing unlabeled data.

Recently, several works have been published regarding BERT compression. \citet{nlptown}, \citet{distilbert} and \citet{Tsai2019SmallAP} showed how BERT can be distilled for training classifiers and sequence taggers with different student models. Our approach differs in that we combine model distillation with a semi-supervised technique for training the compact model using labeled and unlabeled data. We also achieve higher compression ratios. 

Our approach can be used to train compact models for other NLP tasks. For example, by replacing the compact model with other compact models relevant to the downstream task or by replacing or creating an ensemble of teacher models. We present our work-in-progress in the following sections. We intend to release our training method as part of the NLP Architect\footnote{\url{https://github.com/NervanaSystems/nlp-architect}} open source model library.

\section{Approach}

Our approach consists of two models: a \textit{teacher} model which is a pre-trained language model, and a compact (or \textit{student}) model with significantly less parameters, a shallow topology, and which was not pre-trained on any task prior to training on the downstream task.

To transfer knowledge from the teacher model to the compact student model during training we devised a loss function which integrates model distillation and pseudo-labeling of unlabeled examples. Distillation was shown by \citet{Hinton2015DistillingTK} to be an effective method of training compact models by using larger teacher models. Pseudo-labeling is a highly effective semi-supervised technique for bootstrapping using unlabeled data. We use pseudo-labeling to produce a pseudo guess by the teacher model, trusting that the pseudo label is a better guess than randomly choosing a label and by using all available data despite not having annotated NER labels.

\subsection{Models}
In order to provide the compact model with the highest potential for \textit{learning} from a teacher model we opted for a pre-trained language model such as BERT to serve as the teacher model. BERT\cite{Devlin2018BERTPO} is a Transformer-based model that was pre-trained as a masked language model using a large collection of Wikipedia pages and books. \citet{Devlin2018BERTPO} presented two variants of their model, named BERT-base and BERT-large which consist of 110M and 330M parameters, respectively.

The compact, or student, model is shallow, with a simpler topology and is based on the work presented by \citet{Ma2016EndtoendSL}. The compact model consists of convolution layers for extracting character features out of words, a lookup table layer, bi-directional LSTM layers for extracting contextual features and a Softmax or Conditional Random Fields (CRF) as the token classifier. The model has approximately 3M parameters, meaning, it is approximately 36 and 110 times more compact compared to BERT-base and BERT-large. 

\begin{figure}[ht]
\centering
	\includegraphics[scale=0.60]{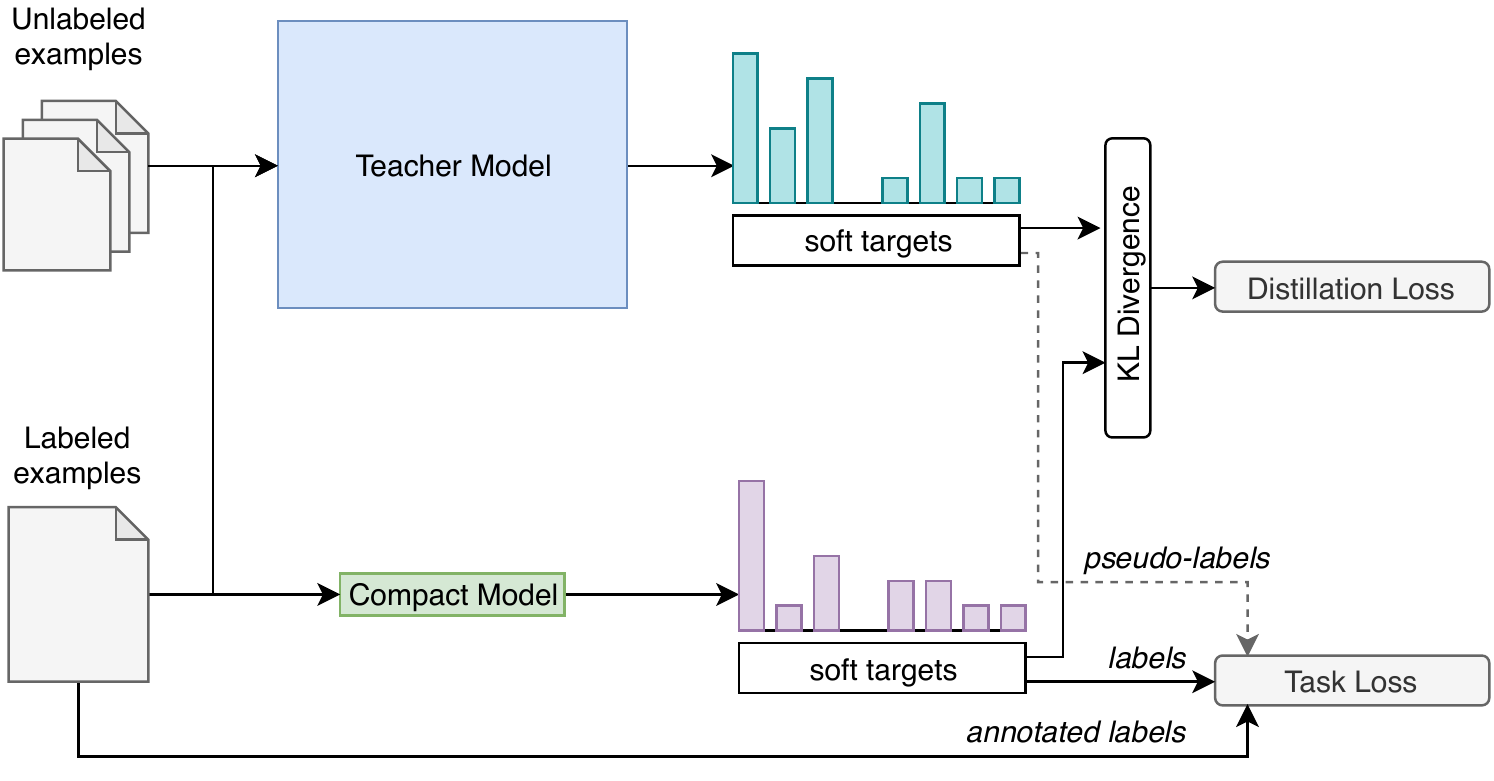}
    \caption{Compact (student) model training process. Teacher and compact models consume examples and produce logits for each example. Distillation loss is calculated by KL-Divergence using both models' logits. The compact task loss is calculated by producing a pseudo-label by the teacher model or by the annotated label that is available in the training set.}
    \label{fig:model}
\end{figure}

\subsection{Training method}

The compact model training process is depicted in Figure \ref{fig:model}. The first step in the training process is to fine-tune the teacher network using the available labeled data to produce a model with the highest possible accuracy on annotated examples. The compact  model is trained using both labeled and unlabeled data in parallel. In each training batch, labeled and unlabeled examples are randomly drawn and the appropriate loss is calculated.
We modified the compact model loss to include three losses: the loss of the downstream task (NER) using the labeled examples, the distillation loss, and the pseudo-labeling loss.
We adapted model distillation from \cite{Hinton2015DistillingTK} and produced the following distillation loss:

$$ L_{distillation} = KL(logits_{compact}/T||logits_{teacher}/T) $$

where $L_{distillation}$ is calculated by using KL-Divergence on the probability distributions of both compact and teacher models. $T$ is the temperature parameter that softens or hardens the logits distribution per class of both models.

We modified the loss of the compact model to include the pseudo-labeling prediction loss. The \textit{pseudo-guess} is produced by applying $\arg\max$ to the logits of the teacher model, and results in the following loss:
$$ L_{task} = \begin{cases}\mathtt{CrossEntropyLoss}(\hat{y}, y) & \text{labeled example}\\\mathtt{CrossEntropyLoss}(\hat{y}, \hat{y}_t) & \text{unlabeled example}\end{cases} $$
where $\hat{y}$ is the entity label class predicted by the compact model and $\hat{y}_t$ is the label predicted by the teacher model.

The final compact model loss is defined by the weighted sum of $L_{task}$ and $L_{distillation}$.


\section{Experiments}

\subsection{Dataset}

In our evaluations we used the English data from the CoNLL 2003 share task \cite{Sang2003IntroductionTT}. The dataset consists of $14987/3466/3684$ samples separated into training, development and test sets, respectively.
We simulate the low-resource scenario by randomly sampling training sets containing $150, 300, 750, 1500$ and $3000$ examples from the training set, which represent approximately 1\%, 2\%, 5\%, 10\% and 20\% of available training examples. The remaining training examples are used as unlabeled training examples. We used the development set to choose the best compact model and report results using the test set.

\subsection{Setup}

Our experimental setup includes BERT-base and BERT-large models \cite{Devlin2018BERTPO} as the teacher models with a token classification head that consists of a fully connected layer and Softmax. We used the pre-trained BERT models provided by the PyTorch-Transformers library\footnote{\url{https://github.com/huggingface/transformers}}. We fine-tuned the teacher models using only the generated labeled training set using default hyperparameters as mentioned in  \citet{Devlin2018BERTPO} for 5 epochs. The compact model was implemented in a fashion similar to the model described in \cite{Ma2016EndtoendSL} and with a similar configuration.
We pre-loaded the embedding layer of the compact models with 100 dimension GloVe \cite{Pennington2014GloveGV} pre-trained word embeddings. The compact models were trained for 20 epochs using batch size 32, with an Adam optimizer with a fixed learning rate of 0.001 and a dropout rate of 0.5.
20 training sets were randomly generated for each training set size, resulting in 100 training sets in all. We report the averaged F1 per training set size using the test set.

\subsection{Compact model accuracy}

\begin{figure}[!tbp]
\centering
  \begin{subfigure}[b]{0.45\textwidth}
    \includegraphics[width=\textwidth]{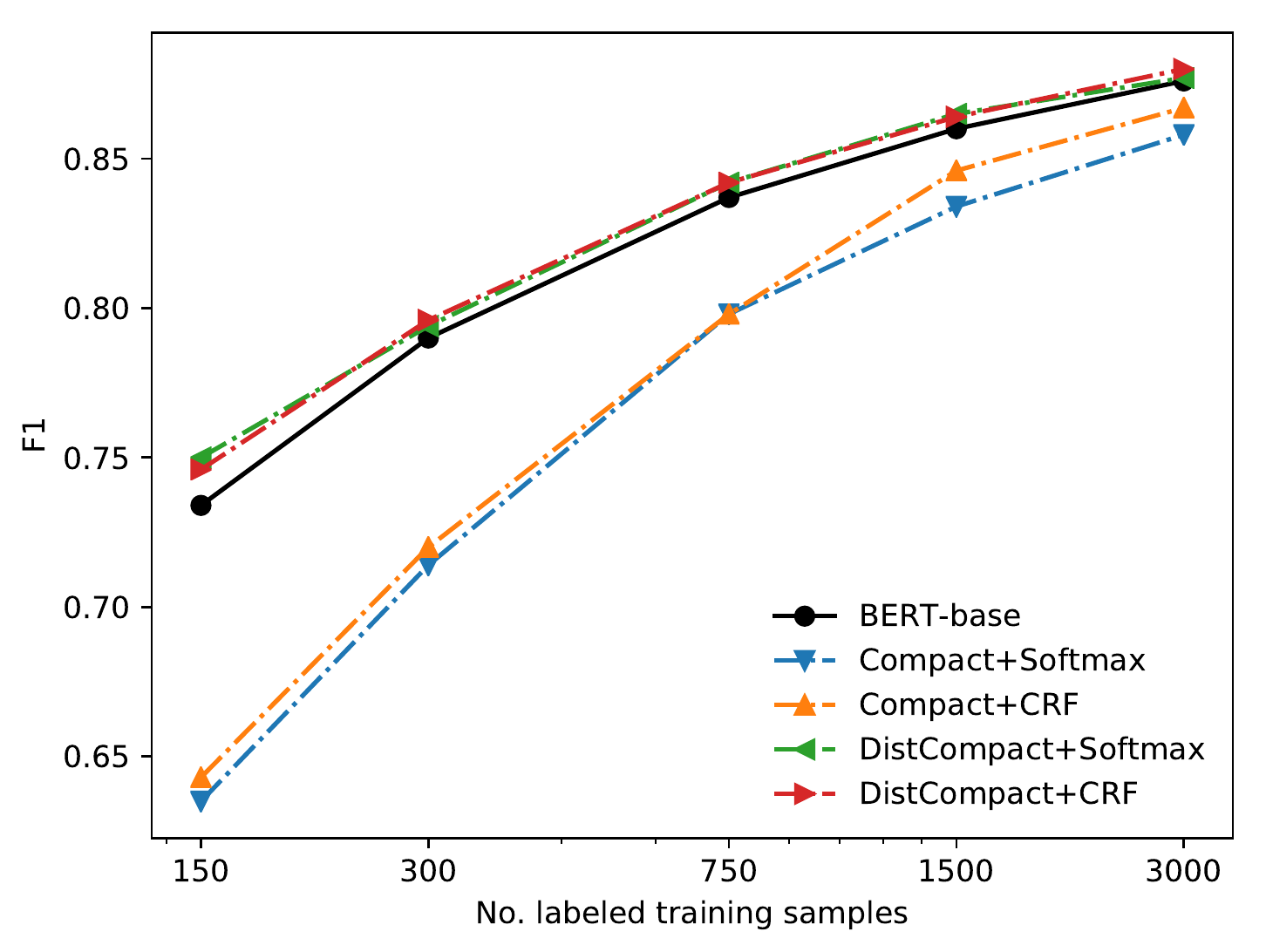}
    \caption{BERT-base as teacher model}
    \label{fig:perf_bert_base}
  \end{subfigure}
  \begin{subfigure}[b]{0.45\textwidth}
    \includegraphics[width=\textwidth]{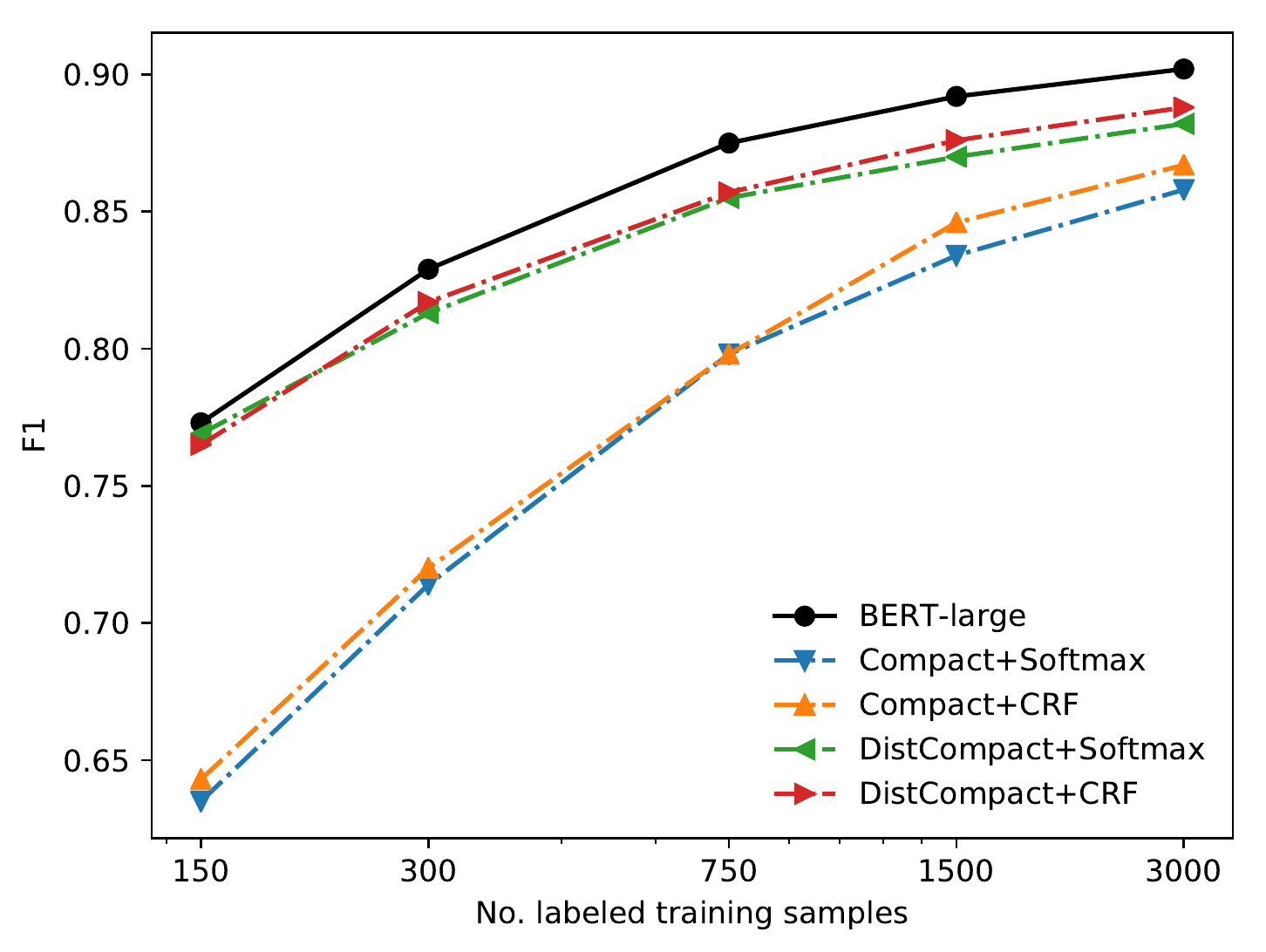}
    \caption{BERT-large as teacher model}
    \label{fig:perf_bert_large}
  \end{subfigure}
  \caption{Model performance (F1) per number of available labeled training samples. \textit{Compact+classifier} denote compact models using only labeled examples, \textit{DistCompact+classifier} denote compact models using distillation and pseudo-labeling training. The compact model classifier is denoted by the suffix (Softmax or CRF). BERT-base was used as the teacher model for training the compact model in (a), was used as the teacher model for training the compact model in (b).}
  \label{fig:accuracy}
\end{figure}

To evaluate the effectiveness of our approach we compared our models with baseline compact models trained using only the available labeled examples and fine-tuned BERT-base and BERT-large models. Figure \ref{fig:accuracy} includes the performance of all models.


As can be seen in Figure \ref{fig:accuracy} the teacher models outperform the baseline compact models in all training set sizes. Figure \ref{fig:accuracy} also shows that the compact models trained using our method consistently outperform the baseline compact models, proving the effectiveness of our approach. Interestingly, compact models trained by our method using less than 300 labeled samples outperform BERT-base and exhibit competitive performance throughout all training set sizes compared with both teacher models. We also learn that as more labeled training data is available, all models start to converge to a similar accuracy with the exception of BERT-large. We also learn that the classifier used by the distilled compact models has almost no effect on the results, as opposed to its effect on the baseline compact models. 

\subsection{Compact model inference speed}

We evaluated the inference speed of the compact models and compared it with both large teacher models on a CPU back-end setup\footnote{Dual Intel® Xeon® Processor E5-2699A v4 @ 2.40GHz; 251GB RAM; OS: Ubuntu 16.04.1 (4.15.0-50-generic); Tensorflow 1.12 and PyTorch 1.0}.
We concatenated all available data and measured total time for completion of 1 pass over all examples. Table \ref{tab:speed} lists our results. As expected, the resulting compact networks show speed-ups of 3.3 to 10.6 for running inference in batch size 1 which is relevant for doing online inference. Batched inference, which is used for information extraction tasks, shows more significant speedups of up to 137.8 when compared with BERT-large. As expected, the simpler Softmax classifiers outperform CRF.

\begin{table}
\caption{Speedup of compact models compared with BERT-base and BERT-large teacher models. Several batch sizes were measured. Speedup is reported per classifier used in the compact model.}
\label{tab:speed}
\centering
\begin{tabular}{@{}lrrrr@{}}
    \toprule
    Batch size & \multicolumn{2}{c}{\textbf{BERT-base}} & \multicolumn{2}{c}{\textbf{BERT-large}} \\
    \multicolumn{1}{r}{} & \multicolumn{1}{c}{\textbf{\begin{tabular}[c]{@{}c@{}}CRF\end{tabular}}} & \multicolumn{1}{c}{\textbf{\begin{tabular}[c]{@{}c@{}}Softmax\end{tabular}}} & \multicolumn{1}{c}{\textbf{\begin{tabular}[c]{@{}c@{}}CRF\end{tabular}}} & \multicolumn{1}{c}{\textbf{\begin{tabular}[c]{@{}c@{}}Softmax\end{tabular}}} \\ \midrule
    1 & 3.3$\times$ & 4.3$\times$ & 8.1$\times$ & 10.6$\times$ \\
    32 & 28.6$\times$ & 33.7$\times$ & 85.2$\times$ & 100.4$\times$ \\
    64 & 40.0$\times$ & 45.2$\times$ & 109.5$\times$ & 123.8$\times$ \\
    128 & 49.9$\times$ & 55.6$\times$ & 123.6$\times$ & 137.8$\times$ \\ \bottomrule
\end{tabular}
\end{table}

\section{Discussion}

Scarcely labeled NLP tasks are hard to tackle, yet are widely common in the industry. Pre-trained language models have been shown to produce state-of-the-art models on such tasks NER, however, the computational and memory requirements make such models difficult to deploy in production. In our initial evaluations we have shown that, for a simulated low-resource NER task, by combining a semi-supervised compression technique such as pseudo-labeling with model distillation we are able to train compact models that can achieve competitive accuracy and superior inference speed compared with larger pre-trained language models. Because our training method is designed to be task independent we intend to evaluate our method on additional NLP tasks, and explore the compression limitations of our approach using other model topologies. More specifically, we plan to explore energy efficient models with the intention of reducing computation (FLOPS) and memory footprint for easy edge-device deployment of multi-model NLP tasks.

\bibliographystyle{abbrvnat}
\bibliography{references}

\end{document}